\documentclass[10pt,twocolumn,letterpaper]{article}

\usepackage[pagenumbers]{cvpr} 

\usepackage{graphicx}
\usepackage{amsmath}
\usepackage{amssymb}
\usepackage{booktabs}
\usepackage{multirow}
\usepackage{adjustbox}
\usepackage[overload]{textcase}
\usepackage{floatrow}
\usepackage{booktabs}
\usepackage{caption}
\usepackage{arydshln}
\usepackage{bbm}
\usepackage{floatrow}
\usepackage[belowskip=-10pt,aboveskip=0pt]{caption}
\usepackage[pagebackref,breaklinks,colorlinks]{hyperref}

\usepackage[capitalize]{cleveref}
\crefname{section}{Sec.}{Secs.}
\Crefname{section}{Section}{Sections}
\Crefname{table}{Table}{Tables}
\crefname{table}{Tab.}{Tabs.}

\def\cvprPaperID{5976} 
\def\confName{CVPR}
\def\confYear{2022}

\begin{document}

\title{ViewCLR: Learning Self-supervised Video Representation \\ for Unseen Viewpoints}

\author{Srijan Das, Michael S. Ryoo\\
Stony Brook University\\
{\tt\small \{srijan.das, mryoo\}@cs.stonybrook.edu}
}
\twocolumn[{%
\renewcommand\twocolumn[1][]{#1}%
\maketitle
\begin{center}
    \centering
    \captionsetup{type=figure}
    \includegraphics[width=1\linewidth]{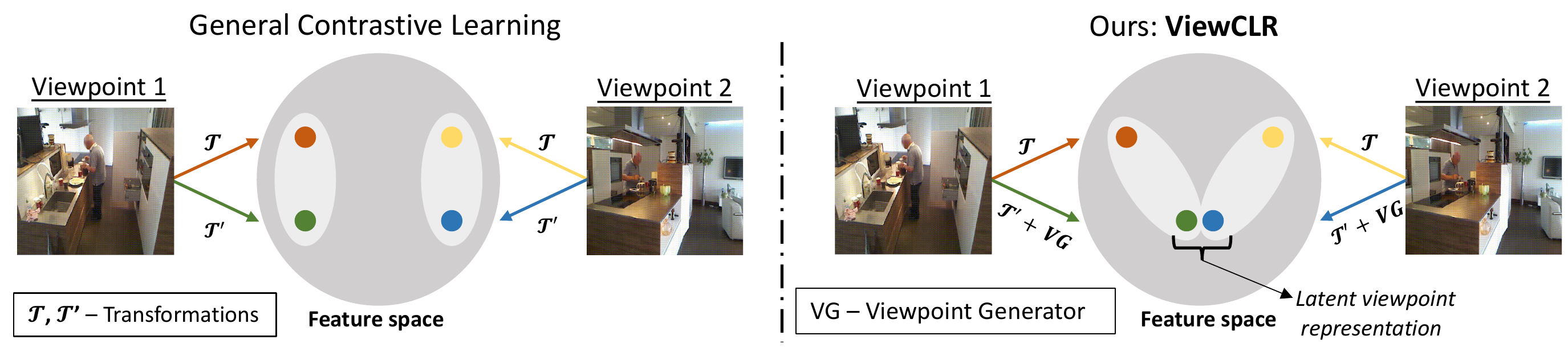}
    \captionof{figure}{An illustration of how contrastive learning with standard data augmentations ($\mathcal{T}$ and $\mathcal{T}'$) embeds two video clips of the same action captured from different camera angles in the feature space versus the feature space representation of ViewCLR. Augmentation through Viewpoint Generator (VG) pulls the mentioned video clips close to one another by generating latent viewpoint representation in the feature space.} \label{title} 
\end{center}%
}]
\begin{abstract}
   Learning self-supervised video representation predominantly focuses on discriminating instances generated from simple data augmentation schemes. However, the learned representation often fails to generalize over unseen camera viewpoints. To this end, we propose \textbf{ViewCLR}, that learns self-supervised video representation invariant to camera viewpoint changes. We introduce a view-generator that can be considered as a learnable augmentation for any self-supervised pre-text tasks, to generate latent viewpoint representation of a video.
   ViewCLR maximizes the similarities between the latent viewpoint representation with its representation from the original viewpoint, enabling the learned video encoder to generalize over unseen camera viewpoints. Experiments on cross-view benchmark datasets including NTU RGB+D dataset show that ViewCLR stands as a state-of-the-art viewpoint invariant self-supervised method. \vspace{-0.2in}
\end{abstract}

\section{Introduction}
\label{sec:intro}
Video understanding has taken a new stride with the advancements of 3D CNNs~\cite{can_spatio-temporal, i3d, C3D}. But one major limitation of CNNs is that they are unable to recognize samples out of training distribution. For instance, if we train an action classifier with videos acquired from one camera viewpoint and test the learned model on videos from a different camera view, the model drastically fails to recognize. This issue persists in a greater extent while learning self-supervised video representation expecting the learned encoder to generalize over a large diversity of viewpoints. 

\noindent Learning self-supervised representation has been very successful with the use of instance discrimination through contrastive learning~\cite{infonce}. The concept of contrastive learning is based on maximizing similarities of positive pairs while minimizing similarities of negative pairs. It is to be noted that the terminology `views' refer to the version of data obtained through data augmentation whereas `viewpoints' refer to the data acquired from different camera angles. Learning invariance to different views is important in contrastive learning. These views can be generated through data augmentation like random cropping, Gaussian blurring, rotating inputs etc. Towards self-supervised video representation, augmentation by tampering the temporal segments in a video is explored to learn invariance across the temporal dimension in videos~\cite{sst1, OPN, sst2, sst3, shufflelearn}. 
However, these pretext tasks and the associated augmentations are not designed to encode viewpoint invariant characteristics to the learned video encoder. 

\noindent In this paper, we focus on learning self-supervised video representation that generalizes over unseen camera viewpoints. We do not aim at designing a new pretext task suitable for a specific downstream scenario but instead propose a module that can be incorporated with the existing self-supervised methods. Several methods have been proposed in the literature to address the challenge of camera view invariant features, mostly using 3D Poses~\cite{2sagcn2019cvpr, msg3d, valstm}. These poses provide geometric information which are robust to camera viewpoint changes. The availability of large scale 3D Poses~\cite{NTU_RGB+D} have facilitated the research community to propose unsupervised skeleton representations~\cite{MS2L, li2021crossclr, sebirenet}. However, the use of 3D Poses are limited to indoor scenarios and most importantly lacks encoding the appearance information. Thus, in this paper, we aim at learning viewpoint invariant features with RGB input in order to generalize self-supervised representations for real-world applications. 

\noindent General contrastive learning methods using standard data augmentation schemes are not explicitly designed to encourage instances pertaining to the same class (for example, similar actions) but different camera viewpoints to pull closer to each other in the feature space as illustrated in Fig~\ref{title}. 
To this end, we propose \textbf{ViewCLR} that provides a learnable augmentation to induce viewpoint changes while learning self-supervised representation. This is achieved by a viewpoint-generator (VG) that learns latent viewpoint representation of a given feature map by imposing the features to follow 3D geometric transformations and projections. However, these transformations and projections are learned by minimizing the contrastive loss. This constraint is achieved by performing a mixup between the features learned by the encoder and the latent viewpoint representation in the manifold space. The outcome is a trained video encoder that takes into account the latent viewpoint representation of the videos, while maximizing its similarities with representation from the original camera viewpoint. As shown in Fig.~\ref{title}, the latent viewpoint representation of the videos enables ViewCLR to pull representations from similar classes but different camera angle close to each other.

\noindent We demonstrate the effectiveness of ViewCLR by evaluating the learned representations for the task of action recognition.
Our experimental analysis shows that ViewCLR significantly improves the action classification performance with regards to generalizing over unseen videos captured from different camera angles. 
On popular datasets with multiple viewpoints like NTU RGB+D and NUCLA, ViewCLR with self-supervised pre-training performs on par with the supervised models pre-trained with huge video datasets. This observation substantiates the importance of learning representations that are invariant to camera angles which is crucial for real-world video analysis.

\section{Background: Neural Projection Layer}\label{bck}
In this section, we recall a recently introduced algorithm Neural Projection Layer (NPL), which learns a latent representation of different camera viewpoint for a given action in supervised settings~\cite{NPL_2021_CVPR}. Our ViewCLR is a spiritual successor of NPL for learning self-supervised viewpoint invariant video representation. 
NPL is derived from the standard 3D geometric camera model used in computer vision. NPL learns a latent 3D representation of actions and its multi-view 2D projections. This is done by imposing the latent action representations to follow 3D geometric transformations and projections, in addition to minimizing the cross-entropy to learn the action labels. First, the feature map $F \in \mathcal{R}^{c \times m \times n}$ which is an intermediate representation of an image of dimension $M \times N$ is fed to a CNN that estimates the 3D space coordinates $p_{x,y}$ for each pixel in $F$. Also, a fully-connected layer is used to estimate the transformation matrices - rotation and translation ($R$ and $t$) for each image in a video. The learned matrices are used to transform a specific camera view to a 3D world coordinate system as $p_{x,y}^w = p_{x,y} \cdot [R^T|R^{T}t]$ for each pixel in $F$. The world 3D representation is then given by:
\begin{equation}
\begin{array}{l}
F^W_{x,y,z} = \sum\limits_{i=0,j=0}^{m,n}(1-|x-p^w_{i,j}[x]|)(1-|y-p^w_{i,j}[y]|) \\   
              \hspace{0.95 in}(1-|z-p^w_{i,j}[z]|)F_{i,j}^{'}
\end{array}
\end{equation}
where $F'$ is obtained by concatenating feature map $F$ and $p_{x,y}$ across channels.

Next, the world feature representation $F^W$ is projected back to 2D. This is done by estimating a camera matrix $K$, which is given by
\begin{equation}
    K = R
\begin{pmatrix}
s_x & 0 & x_0 \\
0 & s_y & y_0 \\
0 & 0 & 1
\end{pmatrix}
\end{equation}
where ($s_x, s_y$) and ($x_0, y_0$) are the scaling factors and the offsets respectively. $R$ here is a $3 \times 3$ camera rotation matrix which is derived from a set of learned parameters. Thus, the 2D projection of the 3D points is estimated as 
\begin{equation}
\begin{array}{l}
F^p_{c,x,y} = \sum\limits_{i=0,j=0}^{m,n}(1-|x-Kp^w_{i,j}[x]|) \\
\hspace{0.9 in} (1-|y-Kp^w_{i,j}[y]|)F_{c,i,j}^{'}
\end{array}
\end{equation}
These frame-level operations are performed across the temporal dimension of a video to compute the viewpoint invariant representation of a video.
In addition to learning the action labels, NPL is constrained over a 3D loss  $\mathcal{L}_{3D}$ as
\begin{equation}\label{3dloss}
    \mathcal{L}_{3D}(V,U) = ||F^W(V) - F^W(U)||_F
\end{equation}
where two videos $U$ and $V$ belong to the same action class. This loss encourages the representations from different viewpoints of the same action result in the same 3D representation. However, learning such viewpoint invariant latent representation with NPL is difficult in the absence of action labels. In ViewCLR, we adopt strategies to learn such latent 3D representation even without the need of human annotated action labels. 
\begin{figure*} 
\scalebox{1}{
   \includegraphics[width=1\linewidth]{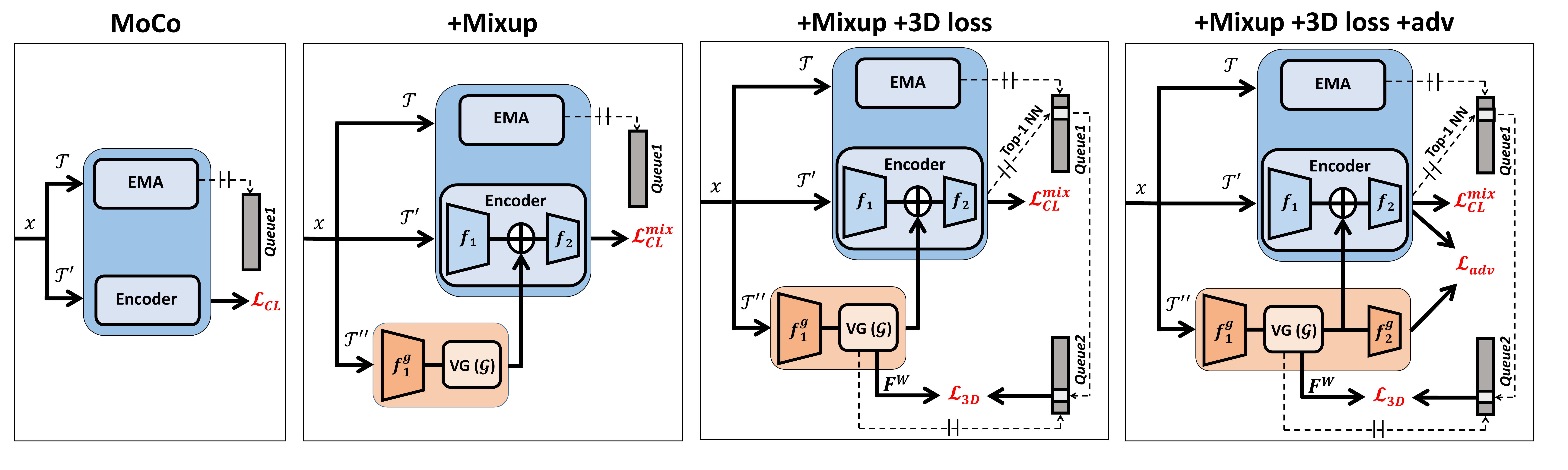} 
   \caption{Illustration of each component in ViewCLR. The input sample $x$ is a video clip. First, the MoCo framework with an encoder and EMA (momentum encoder) is presented at the left. Second, we present ViewCLR with the mixup contrastive loss $\mathcal{L}_{CL}^{\mathrm{mix}}$. Then, the viewpoint-generator (VG) with the 3D loss $\mathcal{L}_{3D}$ is presented. The world 3D representations $F^W$ are encoded in Queue2. Top-1 NN of $f(x)$ in Queue1 is selected from Queue2 for computing the 3D loss with $F^W$. For brevity, we ignore the auto-encoder representation of $F^W$ in this figure. Finally, we present ViewCLR with all its components, including the adversarial loss $\mathcal{L}_{adv}$.}
\label{framework}}
\end{figure*}

\section{ViewCLR}
In this section, we describe our proposed ViewCLR to learn self-supervised video representation such that the learned representation is robust to different viewpoints. For self-supervised representation learning, we use instance discrimination approach proposed in MoCo~\cite{he2019moco}. We remind the readers that the terminology `views' indicates different data views obtained through data augmentation whereas `viewpoints' refer to the videos captured from different camera angles.
\subsection{MoCo}
To formulate, given a video $x$, a set of augmentation transformations $\mathcal{T}$ and $\mathcal{T'}$ is applied on $x$ to generate its two views. Note that these views of the same video are results of standard augmentations and do not involve generating different camera viewpoints. These views of the same video are referred to as positives whereas the augmented $x$ with all other videos are referred to as negatives which are encoded in a dictionary queue referred to as Queue1.
An encoder $f(\cdot)$ and its momentum updated version (EMA) maps the positives and negatives to embedding vectors. Therefore, the InfoNCE loss~\cite{infonce} is formulated as 
\begin{equation}
    \mathcal{L}_{CL} = -\mathrm{log}\frac{\mathrm{exp}(f(x) \cdot {k}_+/\tau)}{ \sum \limits_{i=0}^{\mathcal{N}}\mathrm{exp}(f(x) \cdot {k}_i/\tau)} 
\end{equation}
where $f(x)$ and $k_+$ are the L2-normalized embedding vectors, and $\tau$ is a scaling temperature parameter. The sum is over one positive and $\mathcal{N}$ negative samples in Queue1. For brevity, we loosely use the same notation for the augmented version of input $x$ throughout the paper. This framework as shown in Fig.~\ref{framework}, relies solely on standard transformations $\mathcal{T}$ and $\mathcal{T'}$ to learn discriminative video representation. ViewCLR goes one step beyond to generate latent viewpoint representation of the input videos to learn representation invariant to camera angles. This is achieved by invoking a viewpoint-generator that projects a viewpoint of a video to another arbitrary viewpoint. The question remains, \textit{how do we use such a generator in the MoCo framework}?

\subsection{Viewpoint Generator}
In this section, we describe our proposed viewpoint-generator $\mathcal{G}$ whose working principle is similar to that of NPL but adopted for unsupervised setting. 
First, we justify the design choice of architecture for ViewCLR. Note that we aim at training an encoder with contrastive loss that learns viewpoint invariant representation. The viewpoint-generator is an additional module that can be placed as a block at any intermediate position within the encoder or on top of an encoder. But this design choice would hamper the representation learned by the encoder when we remove the viewpoint-generator for performing downstream tasks. Therefore, ViewCLR introduces another branch in the MoCo framework which consists of an encoder $f^g(\cdot)$ and the viewpoint-generator $\mathcal{G}$. We decompose the encoder with input $x$ as $f^g(x) = f_2^g(f_1^g(x))$ where $f_1^g(\cdot)$ and $f_2^g(\cdot)$ are parts of the encoder $f^g(\cdot)$. We plug in our viewpoint-generator module $\mathcal{G}$ with parameters $\theta_g$ in this stream. 

\noindent In ViewCLR, first an augmentation transformation $\mathcal{T}''$ of the same video $x$ is fed to the partial encoder $f_1^g(\cdot)$. Assuming that we have a viewpoint-generator that can compute latent viewpoint representation of a given feature map, $f_1^g(x)$ is then fed to the viewpoint-generator $\mathcal{G}$. The output of this module $\mathcal{G}(f_1^g(x))$ is a representation of the video projected in an arbitrary  latent viewpoint. The idea is to utilize this representation to train a viewpoint invariant video encoder, so we infuse this feature $\mathcal{G}(f_1^g(x))$ into the MoCo framework by performing a Mixup~\cite{lee2021imix} operation in the manifold space.
\subsubsection{Mixup for infusing the latent viewpoints}
Earlier data mixing strategies~\cite{zhang2018mixup, manifold_mixup, yun2019cutmix, yun2020videomix} have shown that mixing two instances enforces a model to learn discriminative features by providing relevant contextual information. In ViewCLR, we perform mixup between the output feature map of the viewpoint-generator and the intermediate feature map of the encoder $f(\cdot)$. We use this strategy to introduce the output of viewpoint-generator to the MoCo framework.
We perform the mixup in the manifold space as in~\cite{manifold_mixup}.
Let $y_i \in \{0, 1\}^{\mathcal{B}}$ be the virtual labels of the input $x_i$ and its augmented version in a batch, where $y_{i,i} = 1$ and $y_{i,j\neq i} = 0$. Then, the $(\mathcal{N}+1)-$ way discrimination loss for a sample in a batch is:
\begin{equation}
    \mathcal{L}_{CL}(x_i, y_i) = - y_{i,b} \cdot \mathrm{log}\frac{\mathrm{exp}(f(x_i) \cdot k_+/\tau)}{\sum \limits_{j=0}^{\mathcal{N} }\mathrm{exp}(f(x_i) \cdot k_j/\tau)} 
\end{equation}
where $b$ ranges from 1 to $\mathcal{B}$. Thus, the video instances are mixed within a batch for which the loss is defined as:
\begin{equation}
    \mathcal{L}^{\mathrm{mix}}_{CL}((x_i, y_i), (x_r, y_r), \lambda) = \mathcal{L}_{CL}(\mathrm{mix}(x_i, x_r; \lambda), Y) 
    \label{imix_eq}
\end{equation}
where $Y = \lambda y_i + (1-\lambda) y_r$, $\lambda \sim $ Beta($\alpha$, $\alpha$) is a mixing coefficient, $r \sim rand(\mathcal{B})$, and Mix() is the Mixup operator. In ViewCLR, we perform mixup, i.e. simple interpolation of two video cuboids in the feature space such that
\begin{equation}
    Mix(x_i, x_r, \lambda) = \lambda f_1(x) + (1-\lambda)\mathcal{G}(f_1^g(x)) 
\end{equation}
where encoder $f(\cdot)$ is decomposed into $f_1(\cdot)$ and $f_2(\cdot)$. The mixed feature is processed by the partial encoder $f_2(\cdot)$ to optimize the mixed contrastive loss presented in equation~\ref{imix_eq}. This is how, we infuse the latent viewpoint representation $\mathcal{G}(f_1^g(x))$ in the encoder $f(\cdot)$ to minimize the contrastive loss as shown in Fig.~\ref{framework}.

\subsubsection{Latent 3D representation}
\begin{figure}
\scalebox{1}{
   \includegraphics[width=1\linewidth]{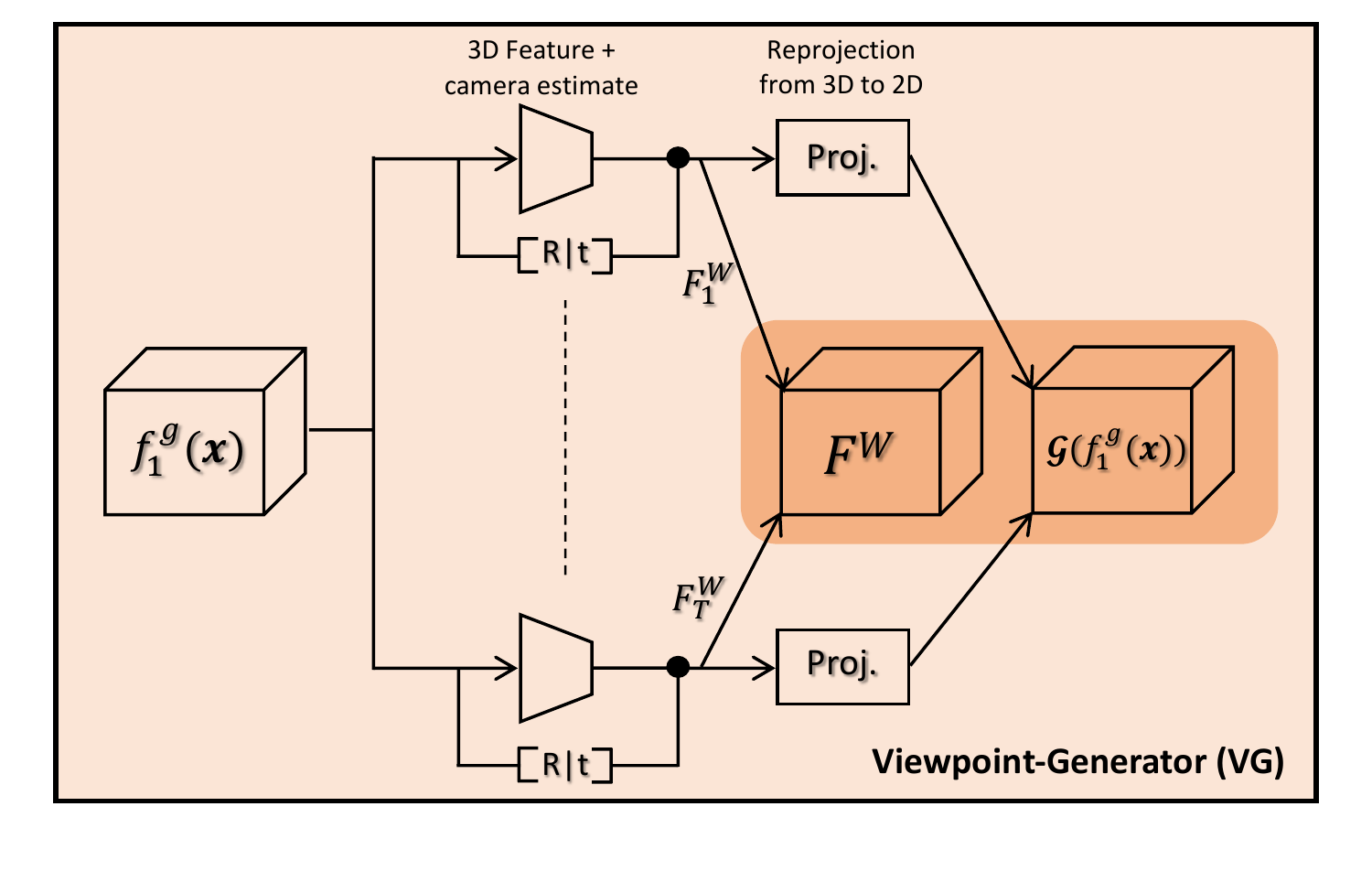} 
   \caption{Learning latent viewpoint representation through Viewpoint-Generator. A CNN computes world feature representation $F^W_j$ at every time-step $j$. The network also learns the transformation and projection matrices to construct a viewpoint invariant 3D representation. Finally, the 3D representation is projected back to 2D space.}
\label{view}} 
\end{figure}

The viewpoint-generator is implemented using NPL and is presented in Fig.~\ref{view}. Here, we extend the notations introduced in section~\ref{bck}. The viewpoint-generator learns the transformation matrices $R$, $t$, and the 3D space coordinates $p_{x,y}$ using the temporal slices of the spatio-temporal feature map $f_1^g(x)$ as input frames. The re-projection of each 3D World representation (for example $F_1^W$) is performed by estimating the camera matrix $K$ within a video. The 2D projected output when combined across all the temporal slices, we obtain the latent viewpoint representation $\mathcal{G}(f_1^g(x))$ for video $x$. 
Different from NPL in~\cite{NPL_2021_CVPR}, the viewpoint-generator here learns the transformation matrices to optimize the mixed contrastive loss $\mathcal{L}_{CL}^{\mathrm{mix}}$. The question remains, \textit{how does the viewpoint-generator learns the world 3D representation}?
Although we do not have the leverage to use action labels, but we propose to mine positive samples from the dictionary queue (Queue1) that encodes history of embeddings. Inspired from the assumptions in~\cite{coclr, li2021crossclr, supportset_2021_ICCV}, we take into account that the nearest neighbor representation of $x$ in Queue1 belongs to the same action category. Consequently on one hand, we obtain the Top-1 nearest neighbor of $f(x)$ in Queue1 such that
\begin{equation}
    NN(f(x), \mathrm{Queue1}) = \mathrm{arg}\min_{q \in \mathrm{Queue1}}||f(x)-q||_2
\end{equation}
On the other hand, we encode the world 3D representation of a video, referred to as $F^W$ in another dictionary queue, namely Queue2. Note that the representation $F^W$ is obtained by combining all the world 3D representation per temporal slices in a video. 
In order to optimize the memory requirement incurred in storing the world 3D representation in Queue2, we use an auto-encoder with a reconstruction loss to squeeze the $(c+3) \times T\times m\times n$ world 3D representation of the video to a $d_{low}$ dimensional embedding vector ($F^W_{inter}$). The details of this auto-encoder is provided in the implementation details.
Thus, the $d_{low}$ dimensional vector after L2-normalization is enqueued to Queue2 while maintaining consistency with the embeddings in Queue1 (see Fig.~\ref{framework}). Now, we reformulate 3D loss $\mathcal{L}_{3D}$ presented in equation~\ref{3dloss} as   
\begin{equation}
    \mathcal{L}_{3D} = ||F^W_{inter}(x) - \mathrm{Queue2}(\mathrm{idx})||_F
\end{equation}
where $\mathrm{idx} = NN(f(x), \mathrm{Queue1})$ represents the index of the Top-1 nearest neighbor 3D world representation in Queue2.

\subsubsection{Adversarial learning of the viewpoint-generator}
Although the viewpoint-generator learns relevant transformations and projection, it might be prone to learning viewpoints similar to the original viewpoint of the input video $x$. Thus, we adopt an adversarial learning to generate different viewpoints as shown in Fig.~\ref{framework}. In contrast to the previous two constraints, here we introduce the partial encoder $f_2^g(\cdot)$. 
The adversarial learning of viewpoint-generator is achieved by using a Gradient Reversal Layer (GRL) on top of $f_2^g(\cdot)$ to maximize the following 
\begin{equation}
    \max_{\theta_g} L_{adv} = \mathbbm{E}_{x \sim P_x}(|f_2^g(\mathcal{G}(f_1^g(x))) - f(x)|_F) 
\end{equation}
where $P_x$ is the data distribution of $x$. This adversarial loss maximizes that the distance between the feature representations $f(x)$ and $f_2^g(\mathcal{G}(f_1^g(x)))$, resulting in generation of complementary camera viewpoints by $\mathcal{G}$.  

\subsection{Training ViewCLR}
Training ViewCLR is quite straightforward. We first train an encoder $f(\cdot)$ using MoCo framework with infoNCE loss for 300 epochs.
Then, we introduce the third stream with partial encoders $f_1^g(\cdot)$ and $f_2^g(\cdot)$ initialized with the weights of $f(\cdot)$ and train them altogether with the summation of mixup contrastive loss $\mathcal{L}^{\mathrm{mix}}_{CL}$, 3D Loss $\mathcal{L}_{3D}$, adversarial loss $\mathcal{L}_{adv}$, and the reconstruction loss for another 200 epochs. This two stage learning enables the encoder to select semantically meaningful Top-1 nearest neighbor, hence benefiting the viewpoint-generator to learn high quality 3D world representation.

\section{Experiments}
In this section, we describe the datasets used in our experimental analysis, implementation details, and evaluation setup. We present ablation studies to illustrate the effectiveness of ViewCLR and also, provide a state-of-the-art comparison.
\subsection{Datasets}
Our dataset choices are based on multi-camera setups in order to provide cross-view evaluation. So, we do not make use of popular datasets like Kinetics-400~\cite{kinetics}, and UCF101~\cite{ucf} to pre-train ViewCLR as the videos in these datasets do not posses the view-point challenges we are addressing in this paper.

\noindent\textbf{NTU RGB+D} (NTU-60 \& NTU-120): NTU-60 is acquired with a Kinect v2 camera and consists of 56k video samples with 60 activity classes. The activities were performed by 40 subjects and recorded from 80 viewpoints. For evaluation, we follow the two standard protocols proposed in~\cite{NTU_RGB+D}: cross-subject (CS) and cross-view (CV). 
NTU-120 is a super-set of NTU-60 adding a lot of new similar actions. NTU-120 dataset contains 114k video clips of 106 distinct subjects performing 120 actions in a laboratory environment with 155 camera views. For evaluation, we follow a cross-subject ($CS_1$) protocol and a cross-setting ($CS_2$) protocol proposed in~\cite{ntu120}.

\noindent\textbf{Northwestern-UCLA Multiview activity 3D Dataset} (NUCLA) is acquired simultaneously by three Kinect v1 cameras. The dataset consists of 1194 video samples with 10 activity classes. The activities were performed by 10 subjects, and recorded from three viewpoints. We performed experiments on N-UCLA using the cross-view (CV) protocol proposed in~\cite{nucla}: we trained our model on samples from two camera views and tested on the samples from the remaining view. For instance, the notation $V_{1,2}^3$ indicates that we trained on samples from view 1 and 2, and tested on samples from view 3. 

\subsection{Implementation Details}
For our experiments with ViewCLR, we use 32 $128 \times 128$ RGB frames as input, at 30 fps. For additional data augmentation, we apply clip-wise consistent random crops, horizontal flips, Gaussian blur and color jittering. We also apply random temporal cropping from the same video as used in~\cite{coclr}. For the encoder backbone $f(\cdot)$, we choose S3D~\cite{s3d} architecture. For the third stream, the viewpoint-generator is plugged after 3 blocks in S3D. So, $f_1^g(\cdot)$ stands for the first 3 S3D blocks and $f_2^g(\cdot)$ stands for 2 S3D blocks. The input to the viewpoint-generator $\mathcal{G}(\cdot)$ is a $480 \times 32 \times 28 \times 28$ spatio-temporal tensor. We use an NPL~\cite{NPL_2021_CVPR} and learn the transformation matrices for each spatial tensor ($480 \times 28 \times 28$). This operation is iterated over 32 temporal slices. The output of NPL from all the time steps are concatenated to obtain $F^W$ (the 3D world feature) and $\mathcal{G}(f_1^g(x))$ for input $x$. 

\noindent The auto-encoder in the viewpoint-generator involves pooling $F^W$ temporally followed by flattening the features (denoted by $F_{inter}$) and fed to two MLPs. The first MLP projects the $(c+3) \times m \times n$ dimensional vector to a lower dimension $d_{low}=128$ and then another MLP to upsample the former to a $(c+3) \times m \times n$ dimensional vector. We invoke a reconstruction loss to minimize the output of the auto-encoder and $F_{inter}$. 

\noindent Hyper-parameters: $\alpha = 1.0$ for mixup, momentum $= 0.999$ for momentum encoder and softmax temperature $\tau = 0.07$. The queue size of MoCo for pre-training is set to 2048. For optimization, we use Adam with $10^{-3}$ learning rate and $10^{-5}$ weight decay. All the experiments are trained on 2 V100 GPUs, with a batch size of 32 videos per GPU.

\subsection{Evaluation setup}
For downstream task, we evaluate the pre-trained ViewCLR models for the task of action classification. We evaluate on (1) \textbf{linear probe} where the entire encoder is frozen and a single linear layer followed by a \textit{softmax} layer is trained with cross-entropy loss, and (2) \textbf{finetune} where the entire encoder along with a linear and \textit{softmax} layer is trained with cross-entropy loss. Note that the encoder $f(\cdot)$ is initialized with the ViewCLR learned weights. More  details for training the downstream action classification framework is provided in the Appendix. 
\begin{table}[t]
  \centering{%
  \scalebox{0.75}{
\caption{Ablation Study of ViewCLR by evaluating on NTU-60 and NUCLA datasets for the task of action classification. The ViewCLR models are trained for 200 epochs after initially training for 300 epochs with infoNCE loss on the training set (protocol CVS) of NTU-60. For a fair comparison, the baseline MoCo model with infoNCE loss, and SkeltonCLR are trained for 500 epochs. The results are evaluated for Linear-probe and finetuned setup for different modalities $\mathcal{M}$ (RGB and Pose). }
\begin{tabular}{c|c|c|ccc|c}
\hline
& \multirow{2}{*}{\textbf{Method}} & \multirow{2}{*}{\textbf{$\mathcal{M}$}} &  \multicolumn{3}{c|}{\textbf{NTU-60}} &  \textbf{NUCLA} \\
\cline{4-7}
 & & &  CVS1 & CVS2 & CVS3  & $V_{3}^{1,2}$ \\
 \hline
{\multirow{6}{*}{\rotatebox{90}{Linear-probe}}} & InfoNCE & R & 28.9 & 20.0 & 20.4 & 37.6 \\ \cline{2-7}
& \textbf{ViewCLR} & R & \textbf{42.5}  & \textbf{32.5} & \textbf{30.6} & \textbf{46.6} \\
& ViewCLR-$\mathcal{L}_{3D}$ & R & 37.5 & 27.1 & 24.9 & 40.3 \\
& ViewCLR-$\mathcal{L}_{Adv}$ & R& 39.2 & 30.3 & 28.1 & 45.8 \\
& ViewCLR-$\mathcal{L}_{Adv}$-$\mathcal{L}_{3D}$& R & 32.6 & 24.9 & 23.8 & 43.9 \\
\cline{2-7}
& SkeletonCLR & P & \textcolor{red}{54.9} & \textcolor{red}{47.0} &  \textcolor{red}{44.3} & - \\
\hline
{\multirow{3}{*}{\rotatebox{90}{Finetune}}} & InfoNCE & R & 82.5 & 74.4 & 73.8 & 81.0 \\
& \textbf{ViewCLR} & R & \textbf{84.2} & \textbf{77.0} & \textbf{75.8} & \textbf{84.6} \\
& SkeletonCLR & P & 81.5 & 69.1 & 65.6 & - \\
\hline
& Supervised & R & 81.4 & 73.4 & 71.0 & \textcolor{red}{87.5} \\
\hline
\end{tabular}\label{ablation}}
} 
\end{table}

\subsection{Ablation Study}
In this section, we empirically show the effectiveness of our viewpoint-generator and the associated loss functions introduced in ViewCLR. Our baseline model is MoCo trained with only infoNCE loss. In Table~\ref{ablation}, we provide the linear-probe and finetuned action classification results on NTU-60 and NUCLA datasets. For our ablation studies, we follow the evaluation protocol proposed in~\cite{varol21_surreact} as it better represents the cross-view challenge. In this protocol, we take only 0$^\circ$ viewpoint from the CS split for training, and we test on the 0◦, 45$^\circ$, 90$^\circ$ views of the cross-subject test split. We call this protocol crossview-subject CVS1, CVS2, and CVS3 respectively. \textit{Our focus is mainly to improve for the unseen and distinct view of 45$^\circ$ and 90$^\circ$}. The models trained on NUCLA are initialized with NTU-60 pre-trained weights.

\noindent In Table~\ref{ablation}, we show that ViewCLR outperforms traditional contrastive model (MoCo) on linear-probe evaluation by a significant margin for both seen (CSV1) and unseen scenarios. The relative improvement on seen camera view-point is 47\% whereas it is upto 62.5\% on unseen camera viewpoint on NTU-60. Next, we provide a full diagnosis of ViewCLR to understand the driving force of this improvement. 

\noindent We remove the 3D Loss that encourages the representation of the videos to project in the same 3D world coordinate system. This model is indicated by ViewCLR-$\mathcal{L}_{3D}$. By default, we also remove the autoencoder and consequently the reconstruction loss from this model. Thus, the viewpoint-generator computes feature map $\mathcal{G}(f_1(x))$ with the adversarial loss
to minimize the mixed contrastive loss. Although the performance of this model is superior to the baseline MoCo model but we show that the absence of the proposed 3D Loss significantly hampers the performance. Thus, the learned representations from the viewpoint-generator highly relies on our proposed 3D Loss. Conversely, we remove the adversarial loss and retain the 3D loss in ViewCLR (referred to as ViewCLR-$\mathcal{L}_{adv}$). This experiment further confirms the effectiveness of our proposed 3D loss. The performance gap of ViewCLR-$\mathcal{L}_{adv}$ model \wrt ViewCLR is owing to the adversarial loss that enables $\mathcal{G}$ to generate latent viewpoint dissimilar to the original viewpoint. Finally, we pre-train ViewCLR by removing the viewpoint-generator in the third stream. So, the third stream only provides feature to perform manifold mixup of the features. This model (ViewCLR-$\mathcal{L}_{adv}$--$\mathcal{L}_{3D}$) is trained only with the mixup contrastive loss $\mathcal{L}_{CL}^{\mathrm{mix}}$. This model further corroborates the effectiveness of the viewpoint-generator. Thus, we show that all the components of ViewCLR along with the proposed losses are instrumental for learning viewpoint invariant video representation. 

\noindent We compare our linear probe results with SkeletonCLR proposed in~\cite{li2021crossclr}. The skeletonCLR result is reproduced in this NTU-60 setting. We note that the linear probe result with 3D Poses as input is superior to that of ViewCLR. This is owing to the properties of poses which are low dimensional informative features. These poses provide geometric information related to the action instances and are robust to viewpoint changes. However, we observe that these poses being low-dimensional input underperforms when the encoder is finetuned. One may argue that the comparison between ViewCLR and skeletonCLR is unfair due to different input modalities but our intent is just to point out the limitation of skeletons which lacks encoding appearance information. Further, we show that the fine-tuned ViewCLR outperforms the InfoNCE based models. For the fully supervised model, the disparity in the performance of NUCLA (87.5\% vs 84.6\%) is because of the supervised NTU pre-training leveraging the human annotations. 

\begin{table}[t]
  \centering{%
  \scalebox{0.85}{
\caption{Comparison of ViewCLR with representative baselines by finetuning the encoders for action classification. The models on NUCLA are pre-trained on NTU-60. Note that the supervised method on NUCLA makes use of the action labels while pre-training on NTU-60.}
\begin{tabular}{|c|cc|c|cc|}
\hline
\multirow{2}{*}{\textbf{Method}}  & \multicolumn{2}{c|}{\textbf{NTU-60}} & \textbf{NUCLA} & \multicolumn{2}{c|}{\textbf{NTU-120}} \\
\cline{2-6}
 &  CS & CV & $V_{3}^{1,2}$ & $CS_1$ & $CS_2$ \\
 \hline
Supervised (S3D)~\cite{s3d} & 85.1 & 86.9 & \textcolor{red}{90.1} & 77.6 & 80.9 \\
MoCo (InfoNCE)~\cite{he2019moco}  & 87.5 & 91.3  & 87.2 & 81.1 & 83.3\\
\hline
\textbf{ViewCLR} & \textbf{89.7} & \textbf{94.1} & \textbf{89.1} & \textbf{84.5} & \textbf{86.2} \\
\hline
\end{tabular}\label{finetune}}
}
\end{table}

\begin{table}[t]
  \centering{%
  \scalebox{1}{
\caption{State-of-the-art linear probe results on NTU-60 for methods using different modalities $\mathcal{M}$.}
\begin{tabular}{|c|c|cc|}
\hline
\multirow{2}{*}{\textbf{Method}}  & \textbf{Modality} & \multicolumn{2}{c|}{\textbf{NTU-60}} \\
\cline{3-4}
 & \textbf{$\mathcal{M}$} &  CS & CV  \\
 \hline
LongTGAN~\cite{LongTGAN} & Poses & 39.1 & 48.1 \\
MS$^2$L~\cite{MS2L} & Poses & 52.6 & - \\
AS-CAL~\cite{ASCAL} & Poses & 58.5 & 64.8 \\
P\&C~\cite{su2020predict} & Poses & 50.7 & \textcolor{red}{76.3} \\
SeBiReNet~\cite{sebirenet} & Poses & - & \textcolor{red}{79.7} \\
Motion decoder~\cite{Motion_decoder_nips} & Flow & \textcolor{red}{77.0} & \textcolor{red}{78.8} \\
3s-CrosSCLR~\cite{li2021crossclr} & Poses & \textcolor{red}{77.8} & \textcolor{red}{83.4} \\
MoCo~\cite{he2019moco} & RGB & 30.5 & 33.5 \\
\hline
\textbf{ViewCLR} & RGB & \textbf{57.0} & \textbf{60.2} \\
\hline
\end{tabular}\label{linear_probe}}
}
\end{table}

\subsection{Comparison to the state-of-the-art}
We clarify that most of our state-of-the-art (SOTA) comparison includes self-supervised approaches using 3D Poses since datasets like NTU RGB+D and NUCLA are popular for skeleton based action recognition. 
In order to provide a fair SOTA, we also present the supervised approaches using (i) models pre-trained on large datasets like Kinetics-400, and (ii) multi-modal (RGB + Poses) information. 

\noindent \textbf{Comparison with representative baselines.} In Table~\ref{finetune}, we present the action classification performance of ViewCLR along with the representative baselines on NTU-60, NUCLA and NTU-120 datasets. Our representative baselines are (i) S3D~\cite{s3d} encoder trained from scratch with randomly initialized weights and (ii) S3D encoder initialized with pre-trained weights learned from MoCo~\cite{he2019moco} framework using InfoNCE loss. Note that all the models on NUCLA are pre-trained on NTU-60. 
We find that ViewCLR outperforms the S3D trained from scratch by a large margin on all the datasets except NUCLA which is pre-trained with action labels. 
Finally, we find that ViewCLR outperforms the self-supervised models trained with traditional InfoNCE loss on all the datasets indicating the effectiveness of ViewCLR for both seen and unseen scenarios.

\noindent \textbf{Linear Probe on NTU-60.} In Table~\ref{linear_probe} and~\ref{SOTA_ntu60}, we present the linear-probing and finetuning SOTA results respectively on NTU-60. While evaluating on linear probing, we observe that ViewCLR significantly outperforms the MoCo model, however lags behind other self-supervised skeleton based methods~\cite{sebirenet, li2021crossclr} and also Motion decoder~\cite{Motion_decoder_nips} leveraging optical flow information. But, we also note that our ViewCLR is superior to all these unsupervised methods when the encoder is fully finetuned in Table~\ref{SOTA_ntu60}. 

\noindent \textbf{Finetuned Results on NTU-60.} ViewCLR with NTU-60 pre-training outperform the models~\cite{glimpse, i3d, NPL_2021_CVPR} which are pre-trained on large datasets. ViewCLR performs on par with the skeleton based models on NTU-60. This is because of the availability of high quality 3D poses on this dataset which may be limited in real-world scenarios. Also, these skeletons do not encode the appearance information while modeling action representation. In order to compare with the models using RGB and Poses, we perform a late fusion of ViewCLR with logits obtained from AGCN~\cite{2sagcn2019cvpr} (Joint stream only). Our ViewCLR + Poses model outperforms all the state-of-the-art results. It also indicates the complementary nature of our RGB based ViewCLR model and the skeleton based AGCN model.
\begin{table}[t]
  \centering{%
  \scalebox{0.85}{
\caption{Comparison to the state-of-the-art finetuned results on NTU-60 for methods using different modalities $\mathcal{M}$ like Poses (P), RGB (R), Depth (D), and Flow (F). The supervised methods presented in this table are pre-trained with action labels in contrast to our unsupervised pre-training. ViewCLR outperforms all the unsupervised methods. Also, ViewCLR when combined with Poses performs on par with the supervised multi-modal methods pre-trained on huge datasets. }
\begin{tabular}{|c|c|c|c|cc|}
\hline
& \multirow{2}{*}{\textbf{Method}}  & \multirow{2}{*}{\textbf{$\mathcal{M}$}} & \textbf{Pre-train}  & \multicolumn{2}{c|}{\textbf{NTU-60}}  \\
\cline{5-6}
& & & \textbf{Dataset} &  CS & CV  \\
\hline
{\multirow{5}{*}{\rotatebox{90}{No pretrain}}} & 2s-AGCN~\cite{2sagcn2019cvpr} & P & $\times$ & 95.1 & 88.5  \\
& MS-G3D~\cite{msg3d} & P & $\times$ & 86.0 & 94.1  \\
& UNIK~\cite{yang2021unik} & P &  $\times$ & 85.1 & 93.6  \\
& CTR-GCN~\cite{ctr_gcn} & P &  $\times$ & 92.4 & 96.8  \\
\cdashline{2-6}
& PEM~\cite{pem} & R+P & $\times$  & 91.7 & 95.2  \\
\cdashline{2-6}
 \hline
{\multirow{5}{*}{\rotatebox{90}{Supervised}}} & Glimpse Cloud~\cite{glimpse} & R & IN1K & 86.6 & 93.2  \\
& I3D~\cite{i3d} & R & IN1K+K400 &  85.5 & 87.3 \\
& NPL~\cite{NPL_2021_CVPR} & R & K400  & - & 93.7  \\
\cdashline{2-6}
& STA~\cite{STA_iccv} & R+P & IN1K+K400  & 92.2 & 94.6 \\
& VPN~\cite{VPN} & R+P & IN1K+K400  & \textcolor{red}{93.5} & 96.2  \\
\hline
{\multirow{6}{*}{\rotatebox{90}{\small{Unsupervised}}}} & MS$^2$L~\cite{MS2L} & P & NTU-60  & 78.6 & -  \\ 
& 3s-CrosSCLR~\cite{li2021crossclr} & P & NTU-60  & 86.2 & 92.5  \\  \cdashline{2-6}
& Motion Decoder~\cite{Motion_decoder_nips} & D & NTU-60  & 68.1 & 63.9 \\
& Colorization~\cite{skeleton_colorization_ICCV} & D & NTU-60 &  88.0 & 94.9  \\
 \cdashline{2-6}
& Motion Decoder~\cite{Motion_decoder_nips} & F & NTU-60  & 80.9 & 83.4  \\ \cdashline{2-6}
& Motion Decoder~\cite{Motion_decoder_nips} & R & NTU-60 &  55.5 & 49.3  \\
\cdashline{2-6}
\hline
{\multirow{2}{*}{\rotatebox{90}{\small{\textbf{Ours}}}}} & \textbf{ViewCLR} & R & NTU-60  &  \textbf{89.7} & \textbf{94.1}  \\
& \textbf{ViewCLR + Poses} & R+P & NTU-60 & \textbf{92.9} & \textbf{97.0} \\
\hline
\end{tabular}\label{SOTA_ntu60}}
} 
\end{table}

\noindent \textbf{Transfer Ability.} In Table~\ref{SOTA_UCLA} and~\ref{SOTA_ntu120}, we present the SOTA results of ViewCLR pre-trained on NTU-60. The action classification accuracy on both the datasets are on par with the SOTA methods which show the generalization capability of ViewCLR.  
Skeleton cloud colorization~\cite{skeleton_colorization_ICCV} outperforms ViewCLR on NUCLA as it takes into account a high dimensional point cloud input which enables this method to address self-occlusions present in this dataset. ViewCLR + Poses performs on par with methods like STA~\cite{STA_iccv} and VPN~\cite{VPN} which are pre-trained on ImageNet + Kinetics-400 (INK), and NTU-60 in contrast to self-supervised NTU-60 ViewCLR pre-training. In Table~\ref{SOTA_ntu120}, we show that the performance of downstream task enhances with increase in the size of pre-trained data. The performance of ViewCLR improves by upto 1.9\% on NTU-120 when the size of the pre-trained data increases from 34K to 55k video samples.

\begin{table}[t]
  \centering{%
  \scalebox{0.85}{
\caption{Comparison to the state-of-the-art finetuned results on NUCLA for methods using different modalities $\mathcal{M}$ like Poses (P), RGB (R), and Depth (D). INK indicates ImageNet1K (IN1K) + K400.}
\begin{tabular}{|c|c|c|c|c|}
\hline
& \multirow{2}{*}{\textbf{Method}}  & \multirow{2}{*}{\textbf{$\mathcal{M}$}} & \textbf{Pre-train} &  \textbf{NUCLA} \\
& & & \textbf{Dataset} & $V_{3}^{1,2}$ \\
\hline
{\multirow{4}{*}{\rotatebox{90}{Super.}}} & Glimpse Cloud~\cite{glimpse} & R & IN1K+NTU-60 & 90.1 \\
& I3D~\cite{i3d} & R & INK+NTU-60 & 88.8\\
\cdashline{2-5}
& STA~\cite{STA_iccv} & R+P & INK+NTU-60 & 92.4\\
& VPN~\cite{VPN} & R+P &INK+NTU-60 & \textcolor{red}{93.5} \\
\hline
{\multirow{3}{*}{\rotatebox{90}{\small{Unsuper.}}}} & MS$^2$L~\cite{MS2L} & P & NTU-60 &  86.8 \\   \cdashline{2-5}
& Motion Decoder~\cite{Motion_decoder_nips} & D & NTU-60 & 62.5 \\
& Colorization~\cite{skeleton_colorization_ICCV} & D & NTU-60 & \textcolor{red}{94.0} \\
\hline
{\multirow{2}{*}{\rotatebox{90}{\small{\textbf{Ours}}}}} & \textbf{ViewCLR} & R & NTU-60  & \textbf{89.1} \\
& \textbf{ViewCLR + Poses} & R+P & NTU-60  & \textbf{92.4} \\
\hline
\end{tabular}\label{SOTA_UCLA}}
} 
\end{table}

\begin{table}[t]
  \centering{%
  \scalebox{0.85}{
\caption{Comparison to the state-of-the-art finetuned results on NTU-120 for methods using different modalities $\mathcal{M}$ like Poses (P), and RGB (R).}
\begin{tabular}{|c|c|c|c|cc|}
\hline
& \multirow{2}{*}{\textbf{Method}}  & \multirow{2}{*}{\textbf{$\mathcal{M}$}} & \textbf{Pre-train}  & \multicolumn{2}{c|}{\textbf{NTU-120}}  \\
\cline{5-6}
& & & \textbf{Dataset} &  $CS_1$ & $CS_2$  \\
\hline
{\multirow{8}{*}{\rotatebox{90}{Supervised}}} & PEM~\cite{pem} & P & $\times$  & 64.6 & 66.9  \\
& 2s-AGCN~\cite{2sagcn2019cvpr} & P & $\times$ & 82.9 & 84.9  \\
 & MS-G3D Net~\cite{msg3d} & P & $\times$  &  86.9 & 88.4 \\
 & CTR-GCN~\cite{ctr_gcn} & P &  $\times$ & \textcolor{red}{88.9} & \textcolor{red}{90.6}  \\
\cdashline{2-6}
& Two-streams~\cite{twostream} & R & $\times$ & 58.5 & 54.8 \\
& I3D~\cite{i3d} & R  & IN1K+K400 & 77.0 & 80.1 \\
\cdashline{2-6}
& Separable STA~\cite{STA_iccv} & R+P & IN1K+K400 & 83.8 & 82.5  \\
& VPN~\cite{VPN} & R+P & IN1K+K400 & 86.3 & 87.8  \\
\hline
{\multirow{4}{*}{\rotatebox{90}{Unsuper.}}}& 3s-CrosSCLR~\cite{li2021crossclr} & P & NTU-120  & 80.5 & 80.4  \\ 
 & \textbf{ViewCLR} & R & NTU-60  & \textbf{82.1} & \textbf{84.3} \\
 & \textbf{ViewCLR} & R & NTU-120  & \textbf{84.5} & \textbf{86.2} \\
& \textbf{ViewCLR + Poses} & R+P & NTU-120  & \textbf{87.1} & \textbf{88.9} \\
\hline
\end{tabular}\label{SOTA_ntu120}}
}
\end{table}

\section{Related work}
 \textbf{Self-supervised video representation.} For learning self-supervised video representation many works have exploited the temporal structure of the videos, such as predicting if frames appear in order, reverse order, shuffled, color-consistency across frames, etc~\cite{sst1, OPN, sst2, sst3, shufflelearn, ss1, ss2, ss3, brave}. On the other hand, some methods have been taking  advantage of the multiple modalities of videos like audio, text, optical flow, etc by designing pretext tasks for their temporal alignment~\cite{out_of_time, cooperative_learning, objects_that_sounds, owens2018audiovisual, evolving_losses, miech20endtoend, look_listen}. Whereas, very less attention is given towards learning viewpoint invariant video representation which is crucial for real-world applications.

\noindent \textbf{View Invariant Action Recognition.} With the advancements in the field of Graph Convolutional Networks~\cite{kipf2016semi} and the availability of abundant 3D Pose data~\cite{NTU_RGB+D}, many works have studied skeleton based action recognition~\cite{stgcn2018aaai,deep-progressive,directed_graph, msaagcn, 2sagcn2019cvpr, msg3d, ctr_gcn}. These skeleton based methods are robust to viewpoint changes due to their extension across depth dimension. Furthermore, to encode appearance information in contrast to the Pose based features, several multi-modal approaches utilizing both RGB and Poses have been proposed in~\cite{rgb+pose_1,rgb+pose_2,Luo_2018_ECCV, Baradel_BMVC, glimpse, STA_iccv, VPN}. 
Recently, several skeleton based self-supervised methods have been proposed in~\cite{MS2L, LongTGAN, ASCAL, su2020predict, sebirenet}. CrosSCLR~\cite{li2021crossclr} performing positive mining across different views (joints, bones, motion) is one of the effective skeleton based self-supervised model till date.
However, these methods utilizing 3D poses are limited to indoor scenarios or availability of high quality poses which is impractical for real-world applications. In contrast, ViewCLR learns viewpoint invariant representation using RGB input only, thus encoding appearance information. Most similar to our work, NPL~\cite{NPL_2021_CVPR} is a geometric based layer to learn 3D viewpoint invariant representation in supervised settings. Different from NPL, ViewCLR can be considered as an augmentation tool for learning self-supervised viewpoint invariant representation. 
\section{Limitations}
In this work, although we aim at learning video representation that generalizes in the wild to different camera viewpoints, our experiments are limited to videos captured in the indoors. This is owing to the scarcity of huge video datasets posing cross-view challenges. On one hand, large datasets like Kinetics~\cite{kinetics} captured from the web mostly contain videos with viewpoint bias. On the other hand, multi-camera datasets like NTU-RGB+D~\cite{NTU_RGB+D}, and MLB~\cite{mlb} are either constrained to indoors or contains less training data.

\section{Conclusion}
We have shown that a complementary viewpoint generation of a video while learning self-supervised video representation can significantly improve the learned representation for downstream action classification task.
We presented ViewCLR that learns latent viewpoint representation of videos through a viewpoint-generator while optimizing the self-supervised contrastive loss. 
Our experiments show the importance of each component of ViewCLR and also confirm its robustness to unseen viewpoints. We believe that ViewCLR is a first step towards generalizing the unsupervised video representation for unseen camera viewpoints and hence, will be a crucial takeaway for the vision community. To facilitate future research, we will release our code and pre-trained representations.

\noindent Future work will explore incorporating ViewCLR on recently proposed successful self-supervised methods~\cite{video_byol, brave}.

\newpage 
\noindent\textbf{Acknowledgements.} 
This work was supported by the National Science Foundation (IIS-2104404). 
We thank Saarthak Kapse and Xiang Li for the great help in preparing the manuscript. 
We also thank members in the Robotics Lab at SBU for valuable discussions.

{\small
\bibliographystyle{ieee_fullname}
\bibliography{egbib}
}

\noindent \textbf{\large{Appendix.}} \\

\noindent \textbf{1. Training/Testing specification for downstream finetuning on NTU60 and NUCLA.} \\
At training, we apply the same data augmentation as in the pre-training stage mentioned in section 4.2., except for Gaussian blurring. The model is trained with similar optimization configuration as in the pre-training stage for 500 epochs. At inference, we perform spatially fully convolutional inference on videos by applying ten crops (center crop and 4 corners with horizontal flipping) and temporally take clips with overlapping moving windows. The final prediction is the average \textit{softmax} scores of all the clips. 

\begin{figure}
\begin{center}
   \includegraphics[width=1\linewidth]{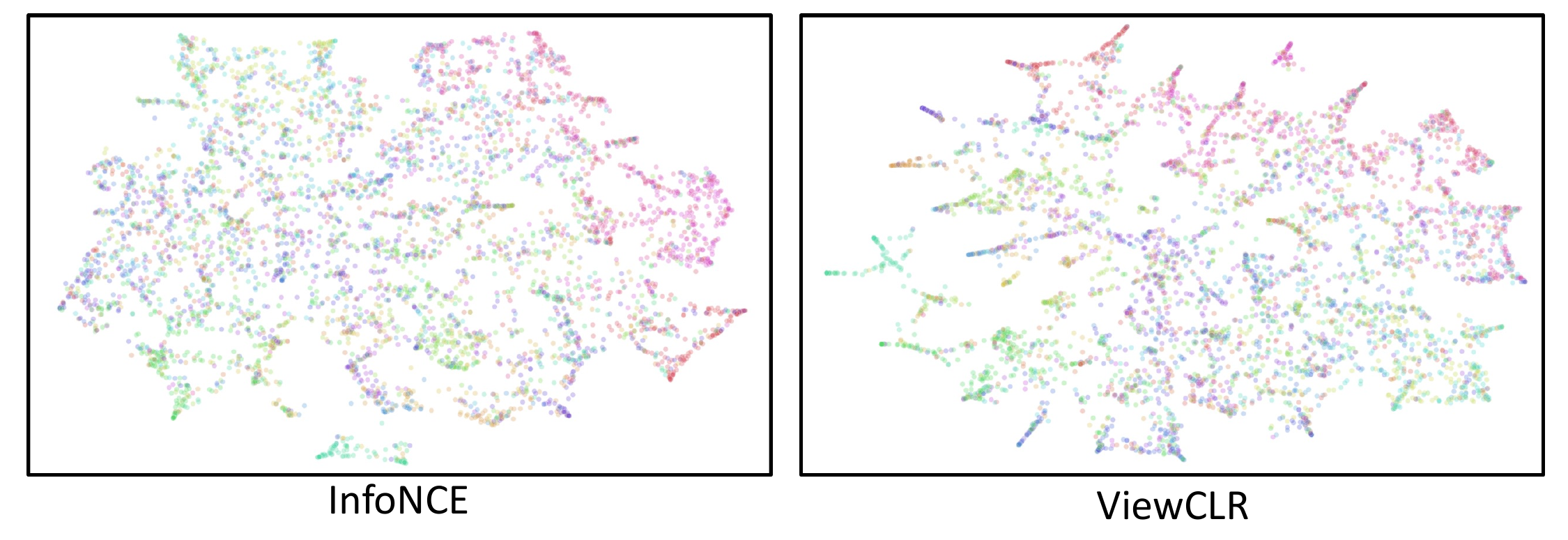} 
\end{center} 
   \caption{t-SNE representation of the test examples captured from unseen viewpoint for InfoNCE (at left) and ViewCLR (at right) models. }
   \label{tsne}
\end{figure}

\noindent \textbf{2. Qualitative Visualization.}\\
In Figure~\ref{tsne}, we provide a t-SNE~\cite{tsne} visualization of the samples captured from unseen camera viewpoints (NTU-60; CVS3 protocol) produced by the InfoNCE and ViewCLR models. These models are trained following the linear probe evaluation, hence the encoders are frozen with the pre-trained weights. We can clearly observe that the ViewCLR model can better discriminate the actions in the feature space in contrast to the traditional InfoNCE model substantiating the importance of the ViewCLR type pre-training.  

\begin{figure*}
\begin{center}
   \includegraphics[width=1\linewidth]{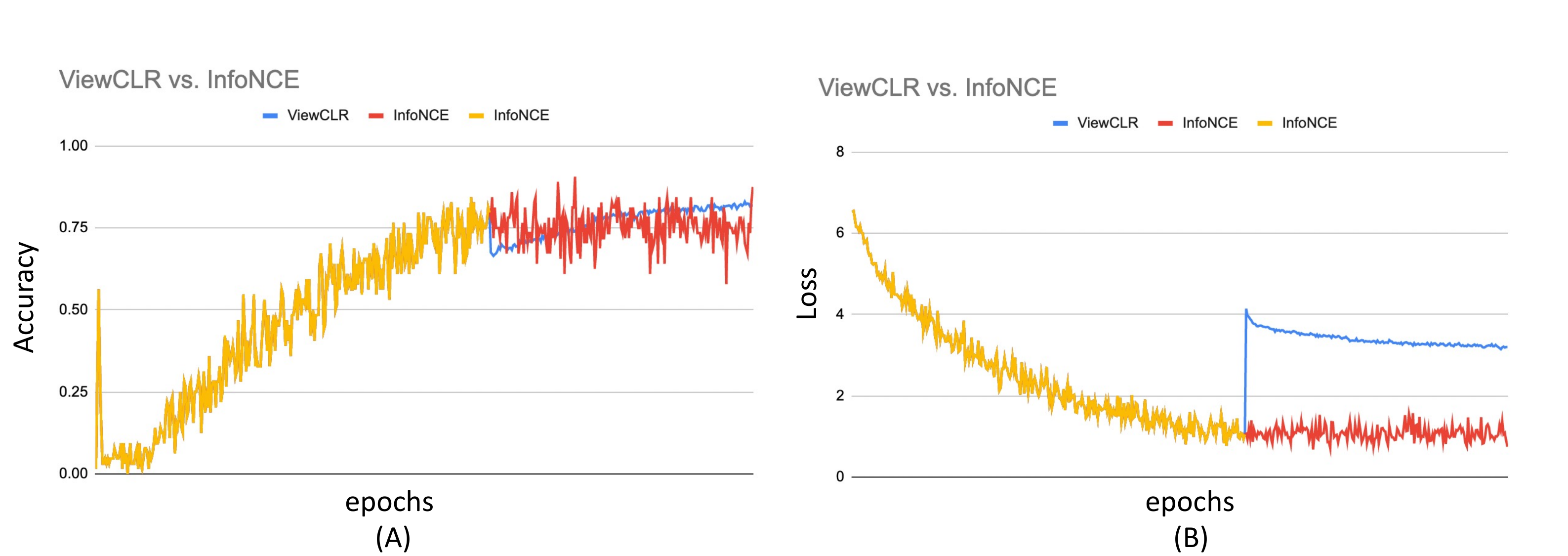} 
\end{center} 
   \caption{Training on NTU-60 dataset. At left, we provide the ($\mathcal{N}+1$)-way accuracy on the pretext task while learning contrastive representation. At right, we present the training loss of ViewCLR and InfoNCE models. ViewCLR models are initially pre-trained with InfoNCE loss for first 300 epochs.}
   \label{graphs}
\end{figure*}
\noindent \textbf{3. Regularization effect of ViewCLR} \\
In Fig.~\ref{graphs}, we provide (A) a plot of ($\mathcal{N}+1$)-way accuracy of the pretext task of the two models, ViewCLR and the MoCo model trained with InfoNCE loss, and (2) the training losses of the two aforementioned models. Note that ViewCLR is trained with InfoNCE loss for first 300 epochs as illustrated in Figure~\ref{graphs}. 
We observe that the ($\mathcal{N}+1$)-way accuracy of the ViewCLR model while training on mixed contrastive loss is lower at times than that of the InfoNCE model (at the left of the figure). However, the performance gain of the ViewCLR model on downstream action classification task tasks shows the regularizing capability of using view-generator. Meanwhile, we observe a disparity between the training losses (at right of the figure) in both the models ViewCLR and InfoNCE. This is owing to the hardness of the pretext task which can be directly correlated with the difficulty of the view-invariant transformation, via the data generator. 
This regularizing capability of ViewCLR is mainly obtained from the mixup strategy introduced in the MoCo model to infuse the view-invariant representation of the videos~\cite{lee2021imix}.
\begin{figure*}

\begin{center}
   \includegraphics[width=1\linewidth]{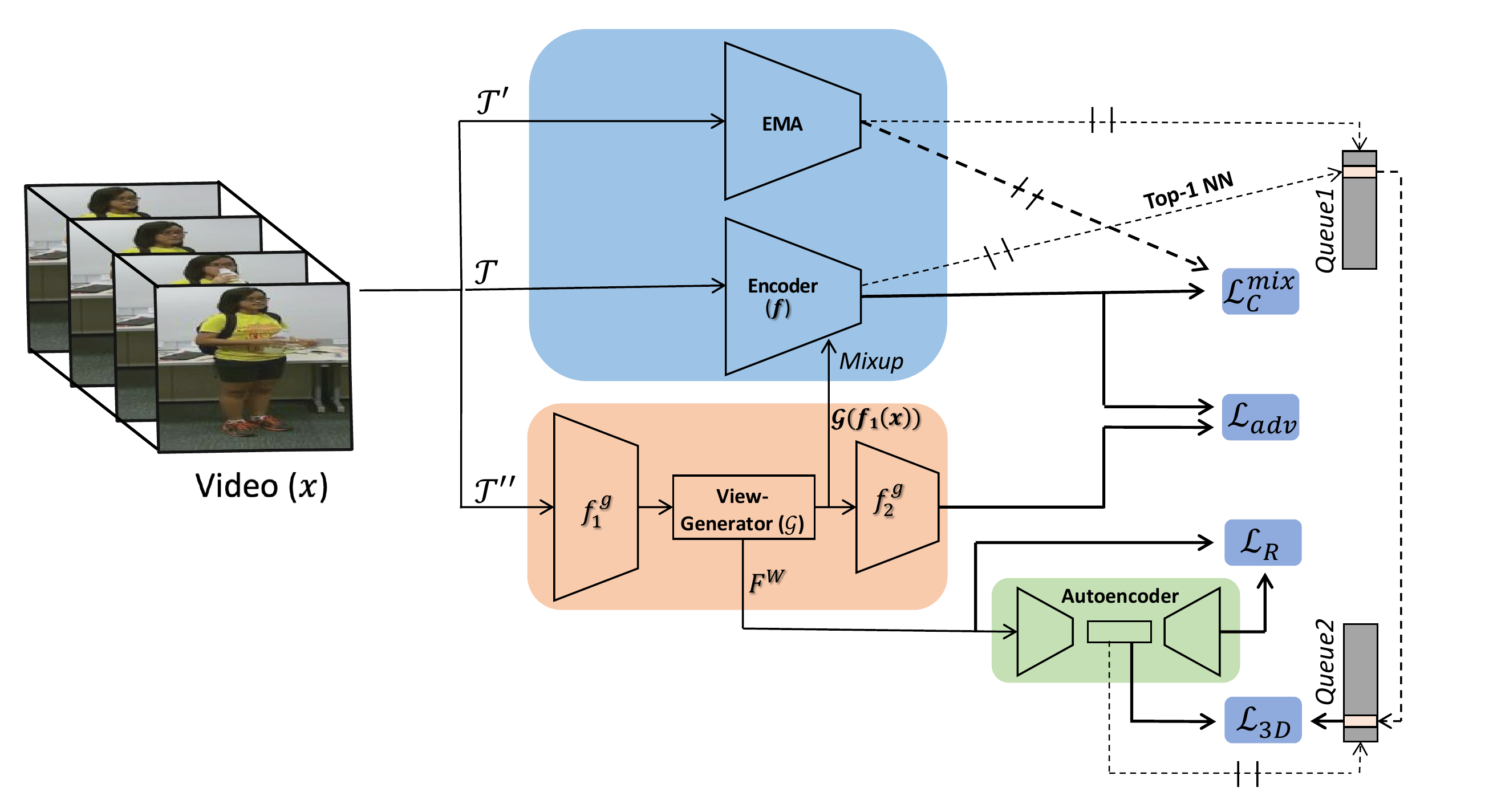} 
\end{center} 
   \caption{Training on UCF101 dataset. At left, we present the training loss of two models, one using video augmentation: STC-mix and the other not. At right, we provide the (K+1)-way accuracy on the pretext task while learning contrastive representation.}
   \label{full_fig}
\end{figure*}

\noindent \textbf{4. Model Architecture.} \\
Figure~\ref{full_fig} provides a full illustration of ViewCLR presented in the main paper. $\mathcal{L}_R$ denotes the reconstruction loss incurred at the output of the view-generator to squeeze the dimension of the world latent feature $F^W$.
\end{document}